\pdfoutput=1

\documentclass[11pt]{article}

\usepackage[final]{acl}

\usepackage{times}
\usepackage{latexsym}

\usepackage[T1]{fontenc}

\usepackage[utf8]{inputenc}
\usepackage{booktabs} 
\usepackage{microtype}

\usepackage{inconsolata}

\usepackage{graphicx}
\usepackage{subfigure}
\usepackage{amsmath}
\usepackage{booktabs}
\usepackage{xcolor}
\usepackage{colortbl}
\usepackage{multirow}
\usepackage{makecell}

\usepackage{CJKutf8}

\title{Cultural Bias Matters: A Cross-Cultural Benchmark Dataset and Sentiment-Enriched Model for Understanding Multimodal Metaphors}

\author{
 \textbf{Senqi Yang\textsuperscript{1}},
 \textbf{Dongyu Zhang\textsuperscript{1}},
 \textbf{Jing Ren\textsuperscript{2}},
 \textbf{Ziqi Xu\textsuperscript{2}},
\\
 \textbf{Xiuzhen Zhang\textsuperscript{2}},
 \textbf{Yiliao Song\textsuperscript{3}},
 \textbf{Hongfei Lin\textsuperscript{1}},
 \textbf{Feng Xia\textsuperscript{2}}
\\
\\
\textsuperscript{1}Dalian University of Technology, China,\\
\textsuperscript{2}RMIT University, Australia,
\textsuperscript{3}The University of Adelaide, Australia\\
\\
\small{
   \textbf{Correspondence:} \href{mailto:zhangdongyu@dlut.edu.cn}{zhangdongyu@dlut.edu.cn}, \href{mailto:ziqi.xu@rmit.edu.au}{ziqi.xu@rmit.edu.au}
 }
}
\begin{document}
\maketitle
\begin{abstract}
Metaphors are pervasive in communication, making them crucial for natural language processing (NLP). Previous research on automatic metaphor processing predominantly relies on training data consisting of English samples, which often reflect Western European or North American biases. This cultural skew can lead to an overestimation of model performance and contributions to NLP progress. However, the impact of cultural bias on metaphor processing, particularly in multimodal contexts, remains largely unexplored. To address this gap, we introduce MultiMM, a Multicultural Multimodal Metaphor dataset designed for cross-cultural studies of metaphor in Chinese and English. MultiMM consists of 8,461 text-image advertisement pairs, each accompanied by fine-grained annotations, providing a deeper understanding of multimodal metaphors beyond a single cultural domain. Additionally, we propose Sentiment-Enriched Metaphor Detection (SEMD), a baseline model that integrates sentiment embeddings to enhance metaphor comprehension across cultural backgrounds. Experimental results validate the effectiveness of SEMD on metaphor detection and sentiment analysis tasks.  We hope this work increases awareness of cultural bias in NLP research and contributes to the development of fairer and more inclusive language models.\footnote{Our dataset and code are available at \url{https://github.com/DUTIR-YSQ/MultiMM}.}
\end{abstract}

\section{Introduction}
\begin{figure}
    \centering
    \subfigure[A metaphor linking paper conservation to a deer.]{
        \includegraphics[width=0.46\linewidth]{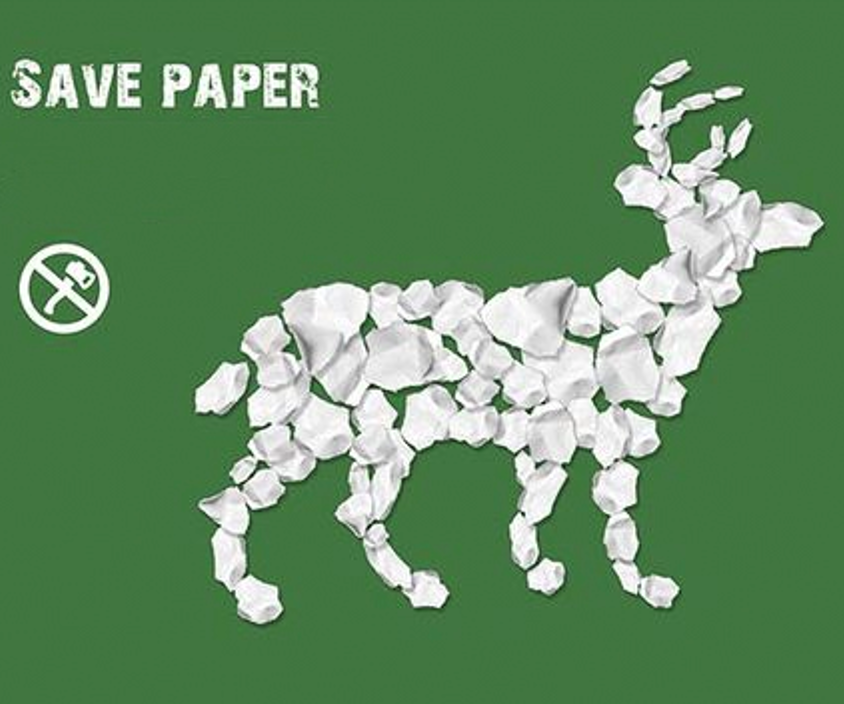}
        \label{Fig:metaphor_example_a}
    }
    \hspace{0.25em}
    \subfigure[Illustrations of common Chinese metaphors.]{
        \includegraphics[width=0.4\linewidth]{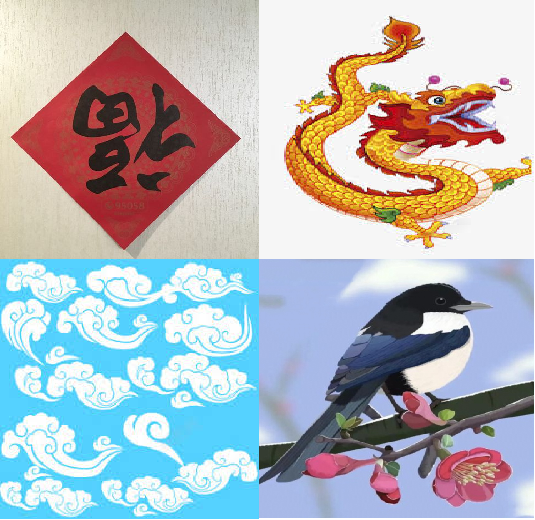}
        \label{Fig:metaphor_example_b}
    }
    \caption{Examples of metaphors across cultures.} 
\end{figure}

Metaphors appear in about one in three sentences and play a pivotal role in human cognition and communication \cite{steen2010method, shutova2010metaphor,hu2023metaphor}. In modern media, multimodal metaphors are more widely used than monomodal ones due to their superior ability to convey vivid and persuasive messages. A multimodal metaphor is a conceptual mapping from one source domain to a target domain, expressed through different combinations of modalities, such as text and image, text and sound, or image and sound \cite{forceville2009multimodal,forceville2021multimodality,zhang2025multimodal}. For example, the text-image combination forming a deer shape in Figure \ref{Fig:metaphor_example_a} creates a metaphorical mapping from the source domain `deer' to the target domain `paper', symbolizing that preserving paper is akin to saving deer.

The shift from monomodal to multimodal metaphors has created a growing need for improved comprehension of multimodal metaphors. However, comprehending them remains a significant challenge for machine-based systems. A key aspect of understanding multimodal metaphors is identifying the underlying mapping between source and target domains while also extracting the conveyed attributes \cite{ su2021multimodal,ge2022explainable}.
Additionally, it involves deciphering implicit messages and recognizing semantic relationships between the source and target domains \cite{ yang2013contextual,zhang2024camel}.

Moreover, implicit messages and semantic relationships can change due to cultural variations, even though conceptual metaphors are deemed universal \cite{kovecses2010metaphor,hong2021cognitive}. For instance, while the mapping from `animal' to `human' is universal, different cultures employ distinct linguistic and visual metaphors to express similar or even identical concepts. The phrase `someone is a dinosaur' means `old-fashioned' in English, while in Chinese, it conveys `ugly'. To describe intoxication, Western cultures use animals such as `newt' or `skunk' (e.g., `as drunk as a skunk'), whereas Chinese culture uses `mud' (e.g., `as drunk as mud'), highlighting a shift in the source domain. Furthermore, symbols of good fortune, such as `fu', `loong', `auspicious clouds', and `picapica' (as shown in Figure~\ref{Fig:metaphor_example_b}), are prevalent in Chinese culture but absent in Western cultures.

Qualitative studies have investigated the interplay between metaphor and culture in fields such as cognitive linguistics \cite{richardson2021cognitive,falck2022procedure,kovaliuk2024connecting}, management \cite{piekkari2020metaphorical,musolff2022war, glaser2024organizations}, and psychology \cite{seering2022metaphors,shokhrukhovna2024metaphorical, xu2024time}. However, the impact of cultural biases on automatic metaphor processing has not been explored in depth. Moreover, the scarcity of metaphor resources from cultures outside Western Europe and North America inevitably leads to cultural bias in current automatic metaphor processing models.

To address this gap, we introduce a benchmark dataset and a sentiment-enriched model for cross-cultural research in multimodal metaphor processing. In summary, this paper makes the following contributions:

\begin{itemize}
    \item We introduce MultiMM, a dataset of 8,461 text-image advertisement pairs, including 4,397 from Eastern cultures and 4,064 from Western cultures. To the best of our knowledge, MultiMM is the first dataset specifically designed for cross-cultural studies in multimodal metaphors.
    
    \item We propose a Sentiment-Enriched Metaphor Detection (SEMD) model that integrates sentiment embeddings to enhance metaphor comprehension across cultural contexts. 
    
    \item We introduce two tasks: metaphor detection and sentiment analysis, to evaluate cross-cultural multimodal metaphor understanding. The evaluation results of eight textual, three visual, and seven multimodal baselines demonstrate the significant impact of cultural bias on metaphor processing.
    
    \item We present a new perspective on cultural bias in multimodal metaphor processing. Through the identification and analysis of these biases, our work aims to inspire future research that promotes awareness of cultural disparities and supports the creation of fairer, more inclusive NLP systems.
\end{itemize}

\section{Related Work}
\subsection{Multimodal Metaphor Datasets}
Most existing metaphor datasets are limited to text-only formats \cite{birke2006clustering, steen2010method, mohammad2016metaphor, chen-etal-2023-chinese, tong-etal-2024-metaphor}, resulting in a significant gap in multimodal metaphor research. To date, few multimodal metaphor datasets exist, and most originate from North American and Western European contexts \cite{shutova2016black,zhang2021multimet}\footnote{\url{https://github.com/DUTIR-YSQ/MultiMET}}, with limited representation of other cultural perspectives \cite{xu2022met,lu2025emometa,zhang2023multicmet}\footnote{\url{https://github.com/DUTIR-YSQ/MultiCMET}}. In contrast, MultiMM provides diverse annotations for metaphor understanding across Eastern and Western cultures, addressing the limitations of existing benchmarks.

\subsection{Metaphor Understanding}
Early studies on metaphor comprehension in textual data primarily relied on manually crafted knowledge \cite{mason2004cormet}. Subsequent research explored distributional clustering \cite{shutova2013statistical} and unsupervised learning approaches \cite{shutova2017multilingual, mao2018word}. More recently, deep learning models have been employed to improve metaphor comprehension. With the advent of large language models (LLMs), further progress has been made in this domain \cite{liu2020metaphor, choi2021melbert, stowe-etal-2021-metaphor, GeCambria2022, aghazadeh-etal-2022-metaphors, wachowiak-gromann-2023-gpt, tian-etal-2024-bridging}. While some studies have explored multimodal metaphor processing, they have primarily focused on extracting and integrating textual and visual features \cite{shutova2016black, kehat2020improving, su2021enhanced, xu2024exploring, xu2024generating}. Distinct from prior work, our model SEMD incorporates a cultural perspective by leveraging sentiment information, which is a universally recognized feature for multimodal metaphor understanding.

\section{The MultiMM Dataset}
\subsection{Data Collection and Filter}
MultiMM aims to provide a cross-cultural, labeled dataset for automated multimodal metaphor comprehension. Given that metaphorical content is primarily textual or visual \cite{steen2010method,akula2023metaclue}, we collect data from commercial and public service advertisements incorporating both textual and visual elements to enable the analysis of visual and linguistic features for multimodal metaphor understanding. A statistical overview of MultiMM is presented in Table \ref{tab1}.

\begin{table}[t]
    \small 
    \centering
    \setlength{\tabcolsep}{4pt}
    \begin{tabular}{@{}cccc@{}} 
        \toprule
        \textbf{Item} & \textbf{CN} & \textbf{EN} & \textbf{Total} \\ 
        \midrule
        Total & 4,397 & 4,064 & 8,461 \\ 
        Metaphorical & 2,583 & 2,189 & 4,772 \\ 
        Literal & 1,814 & 1,875 & 3,689 \\ 
        \midrule
        Total Words & 145,312 & 68,189 & 213,501 \\ 
        Average Words & 33 & 15 & 24 \\ 
        \midrule
        Training Set Size & 3,517 & 3,251 & 6,768 \\ 
        Validation Set Size & 440 & 406 & 846 \\ 
        Test Set Size & 440 & 407 & 847 \\ 
        \bottomrule
    \end{tabular}
    \caption{Statistical overview of MultiMM. CN refers to Chinese data, while EN refers to English data.}
    \label{tab1}
\end{table}

\begin{CJK*}{UTF8}{gbsn} 

\noindent\textbf{Chinese Advertisement Collection}. MultiMM contains 4,397 Chinese samples, which native Chinese-speaking researchers collect by searching for Chinese keywords via Baidu~\footnote{\url{https://www.baidu.com/}}. We compile a set of keywords related to `advertisement' and `metaphor', encompassing three main categories: everyday products (e.g., 手机~[mobile], 汽车~[car]), public service topics (e.g., 吸烟~[smoking], 欺凌~[bullying]), and metaphorically relevant linguistic concepts (e.g., 愤怒~[anger], 颜色~[color]).

We also refer to the Master Metaphor List \cite{lakoff1994master} to select target and source domains in conceptual metaphors, using keywords such as 变化~[change], 情感~[sentiment], 人们~[people], and 信念~[beliefs]. In addition, we collect potential Chinese metaphorical samples that contain both images and text from Chinese commercial advertisements released in 2021, following the iFlytek Advertising Picture Classification Competition~\footnote{\url{https://aistudio.baidu.com/aistudio/datasetdetail/102279}}.

\end{CJK*}

\noindent\textbf{English Advertisement Collection}. The 4,064 English samples come from a public dataset containing product and public service advertisements with both images and text \cite{ye2021interpreting}. We further clean the data by: (1) removing duplicate images based on their MD5 encoding \cite{rivest1992md5}; (2) manually removing images that are not advertisements; (3) removing images that are blurry or smaller than 350 × 350 pixels; (4) extracting the embedded text using the optical character recognition (OCR) technique and manually correcting the results.

\subsection{Annotation Model}

\begin{figure}[t]
  \centering
  \includegraphics[width=0.8\linewidth]{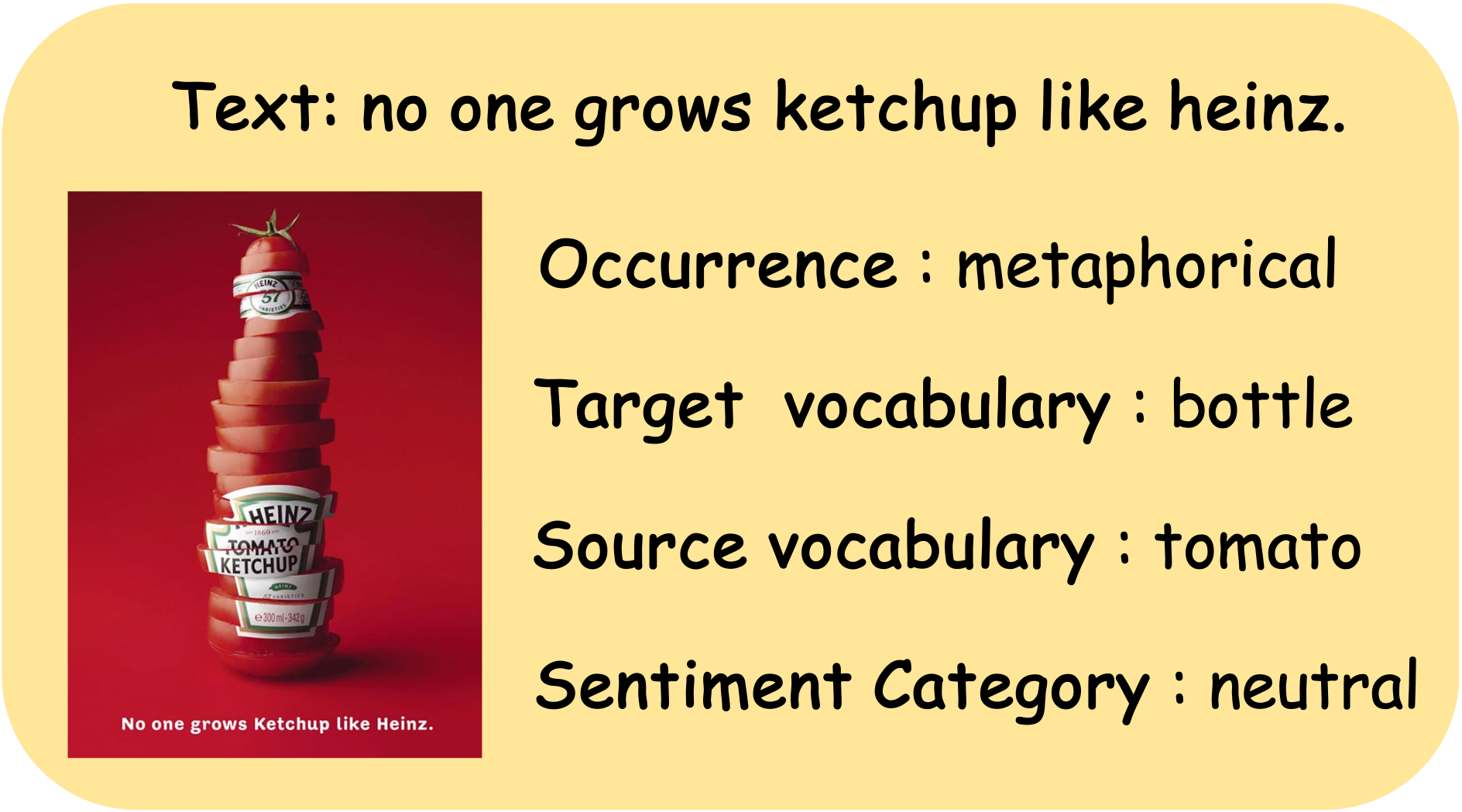}
  \caption{Example of an advertisement with annotations. This figure shows an English advertisement in which stacked tomatoes visually form a ketchup bottle, accompanied by the text ‘No one grows ketchup like Heinz’. The example is annotated as metaphorical, with ‘tomato’ as the source vocabulary, ‘bottle’ as the target vocabulary, and the sentiment labeled as neutral.}
  \label{fig:figure_2}
\end{figure}

We annotate the text-image advertisement pairs based on the following criteria:  (1) the occurrence of metaphors (literal or metaphorical);  (2) target and source domain relations, including target/source vocabulary in text and verbalized target/source vocabulary in images;  (3) sentiment category, classified as negative, neutral, or positive.  

The annotation model is defined as AnnotationModel = (Occurrence, Target, Source, SentimentCategory). Figure \ref{fig:figure_2} illustrates an example of an advertisement with annotations. Additionally, we provide detailed annotation guidelines via the link in the Abstract.

\subsection{Data Annotation}
\textbf{Metaphorical or literal}. Following MultiMET \cite{zhang2021multimet}, we identify metaphorical and literal expressions in both verbal and visual forms. For each identified metaphor, annotators specify its domains: textual metaphors derive source and target domains from the original words, while visual metaphors require annotators to verbalize domain words inferred from the image. Adopting the approach of~\citet{vsorm2018towards}, annotators detect metaphorical text-image pairs by identifying incongruous elements and explaining a non-reversible `A is B' identity relation, which signifies two domains expressed across different modalities.

\noindent\textbf{Sentiment categories}. Understanding metaphors involves identifying domain mappings and analyzing linguistic properties, such as sentiment. Metaphors often have a stronger emotional impact than literal expressions \cite{mohammad2016metaphor,schnepf2022s}. To explore this, we annotated sentiment in MultiMM as negative, neutral, or positive to compare its impact on metaphorical and literal expressions across multicultural and multimodal contexts.

\subsection{Annotation Process and Quality Control}
We use an expert-based approach to annotate data with eight researchers: five Chinese native speakers (grouped into 2-2-1) annotate Chinese data, and three English native speakers (grouped into 2-1) annotate English data. The two-expert groups annotate data, while the one-expert groups resolve disagreements. If agreement is not reached, all experts discuss to finalize the annotation. To ensure quality, we implement strict annotation standards and provide detailed documentation with instructions, examples, and cautions. Training sessions are conducted before each annotation round, and all training materials and documents are continuously adjusted to address new challenges. Additionally, we pre-label a small dataset to identify and resolve potential issues early in the process.

Two strategies are used to mitigate cultural bias during annotation: Diverse Annotator Selection and Feedback Mechanisms. Diverse Annotator Selection ensures that annotators have multicultural backgrounds. For example, two of our five Chinese expert annotators have lived and studied in Western countries for over six years, while the other three have resided there for more than two years. Similarly, all three English annotators have over two years of experience living and working in China or other Eastern countries. This helps minimize the influence of any single cultural perspective. The Feedback Mechanisms allow annotators to share insights and flag potential cultural biases during the annotation process. The open communication channel also ensures prompt issue resolution and improves annotation quality.

To measure the consistency of the classification, we use the Fleiss' Kappa score ($\kappa$) \cite{fleiss1971measuring}, a statistical metric that evaluates the agreement among three or more raters while accounting for chance agreement. The Fleiss' kappa score ranges from -1 to 1, where $\kappa > 0.6$ is considered substantial agreement, while $\kappa > 0.8$ indicates near-perfect agreement. The score for metaphor identification is $\kappa = 0.73$, for target domain category identification is $\kappa = 0.70$, for source domain category identification is $\kappa = 0.66$, and for sentiment category identification is $\kappa = 0.82$. These results indicate that the annotation process is reliable.

\begin{figure}[t]
    \centering
    \includegraphics[width=0.8\linewidth]{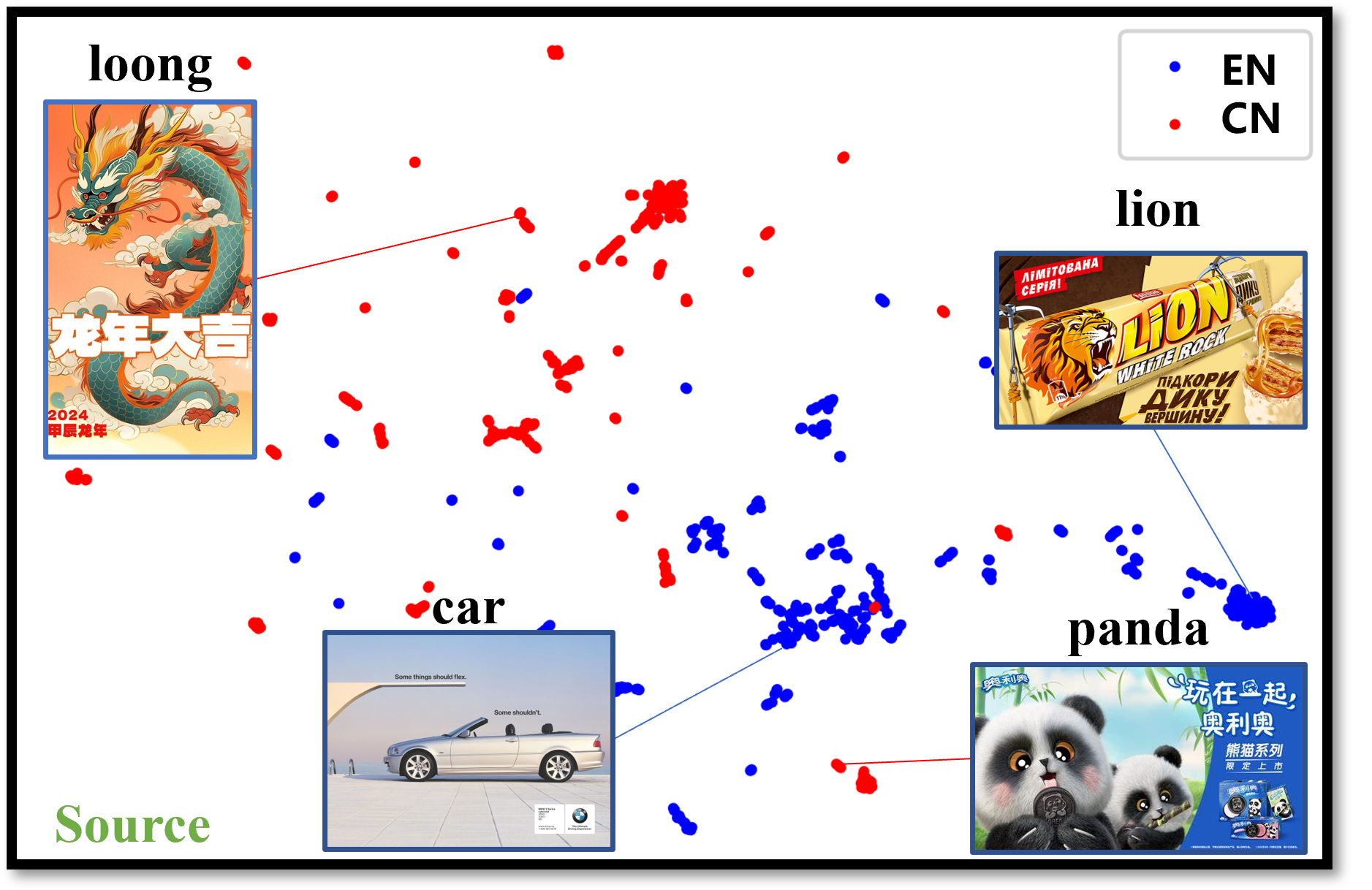}
    \label{Fig:figure_3_b_ziqi}
    \caption{The distribution of source domain vocabulary in Chinese (CN) and English (EN) advertisements.} 
    \label{fig:figure_3}
\end{figure}

\section{Data Analysis}
\subsection{Metaphor Distribution}
As shown in Figure \ref{fig:figure_3}, the distribution of source domain vocabulary in English and Chinese samples differs significantly. English advertisements often utilize concrete, universally recognized source vocabulary to evoke immediate associations. Animals such as lions and eagles symbolize strength and freedom, while vehicles like rockets and sports cars represent innovation and luxury. In contrast, Chinese advertisements frequently draw on source vocabulary that reflects traditional symbolism and societal values. Animals such as loong and pandas symbolize power and national pride, while natural elements like bamboo and lotus flowers represent resilience and purity.

Despite these cultural differences, certain source-target pairs appear in both English and Chinese advertisements, reflecting universal human experiences or the homogenizing effects of globalization. For instance, the cheetah is frequently associated with automotive products in both cultures to emphasize speed. Similarly, water often symbolizes cleaning agents or bottled beverages to convey purity, though its visual representation may vary across cultures. However, many metaphorical relationships remain culture-specific, shaped by distinct social and linguistic contexts. Natural elements such as fire exemplify this divergence: in English advertisements, fire is often linked to luxury perfumes to evoke passion, as seen in campaigns featuring fiery motifs, whereas, in Chinese contexts, fire is rarely used for such purposes.

\begin{figure}[t]
  \centering
  \includegraphics[width=0.99\linewidth]{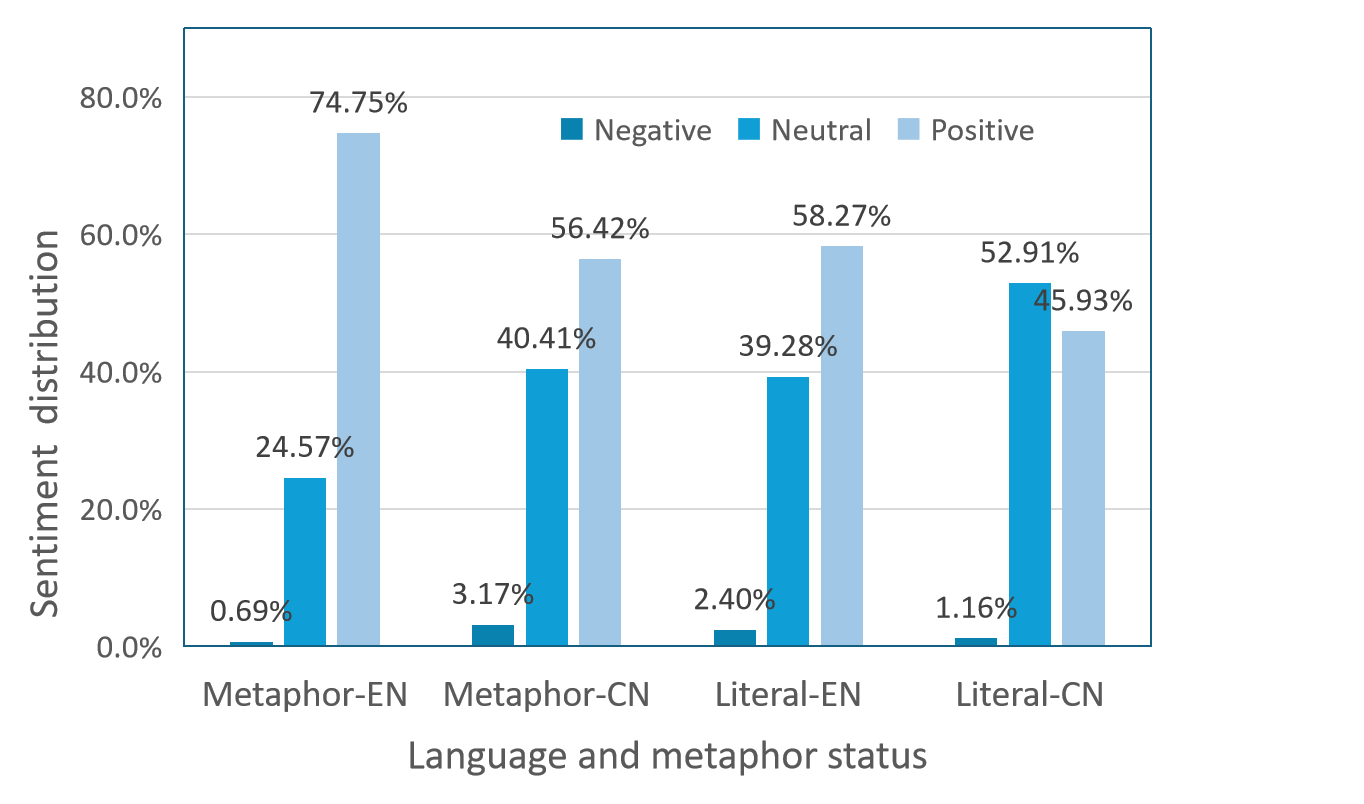}
  \caption{Sentiment category distribution in metaphorical and literal data from Chinese (CN) and English (EN) advertisements.}
  \label{fig:figure_4}
\end{figure}

\subsection{Sentiment Category Distribution}
Figure \ref{fig:figure_4} compares sentiment distributions in English and Chinese metaphorical and literal advertisement data. In metaphorical advertisements, English data exhibits a pronounced emphasis on positive sentiment (74.75\%), with neutral sentiment at 24.57\% and no negative sentiment (0.69\%). This contrasts with Chinese metaphorical advertisements, where positive sentiment remains dominant (56.42\%) but is tempered by a higher level of neutrality (40.41\%) and a small presence of negativity (3.17\%). These patterns suggest that while metaphors in both languages are leveraged to evoke positivity, English advertisements employ figurative language more assertively to amplify optimism, whereas Chinese advertisements strike a balance between positivity and neutrality. The complete absence of negative sentiment in English metaphorical advertisements may reflect cultural taboos against associating metaphors with negativity in advertising. In contrast, the minimal negative sentiment in Chinese metaphorical advertisements could serve as a strategic rhetorical device, such as juxtaposing challenges with solutions to enhance persuasive impact.

In literal contexts, sentiment dynamics shift significantly. In English advertisements, positive sentiment declines sharply to 58.27\%, while neutral sentiment rises to 39.28\%, and negative sentiment emerges at 2.40\%. Chinese literal advertisements exhibit an even more pronounced shift: neutral sentiment dominates (52.91\%), surpassing positive sentiment (45.93\%), with negative sentiment remaining minimal (1.16\%). The data confirms that metaphors consistently convey richer emotional resonance than literal expressions across both languages. These findings align with prior studies \cite{mohammad2016metaphor,stanley2021collective}, which argue that metaphors amplify emotional engagement by activating deeper cognitive and affective associations.

\begin{figure}[t]
  \centering
  \includegraphics[width=0.9\linewidth]{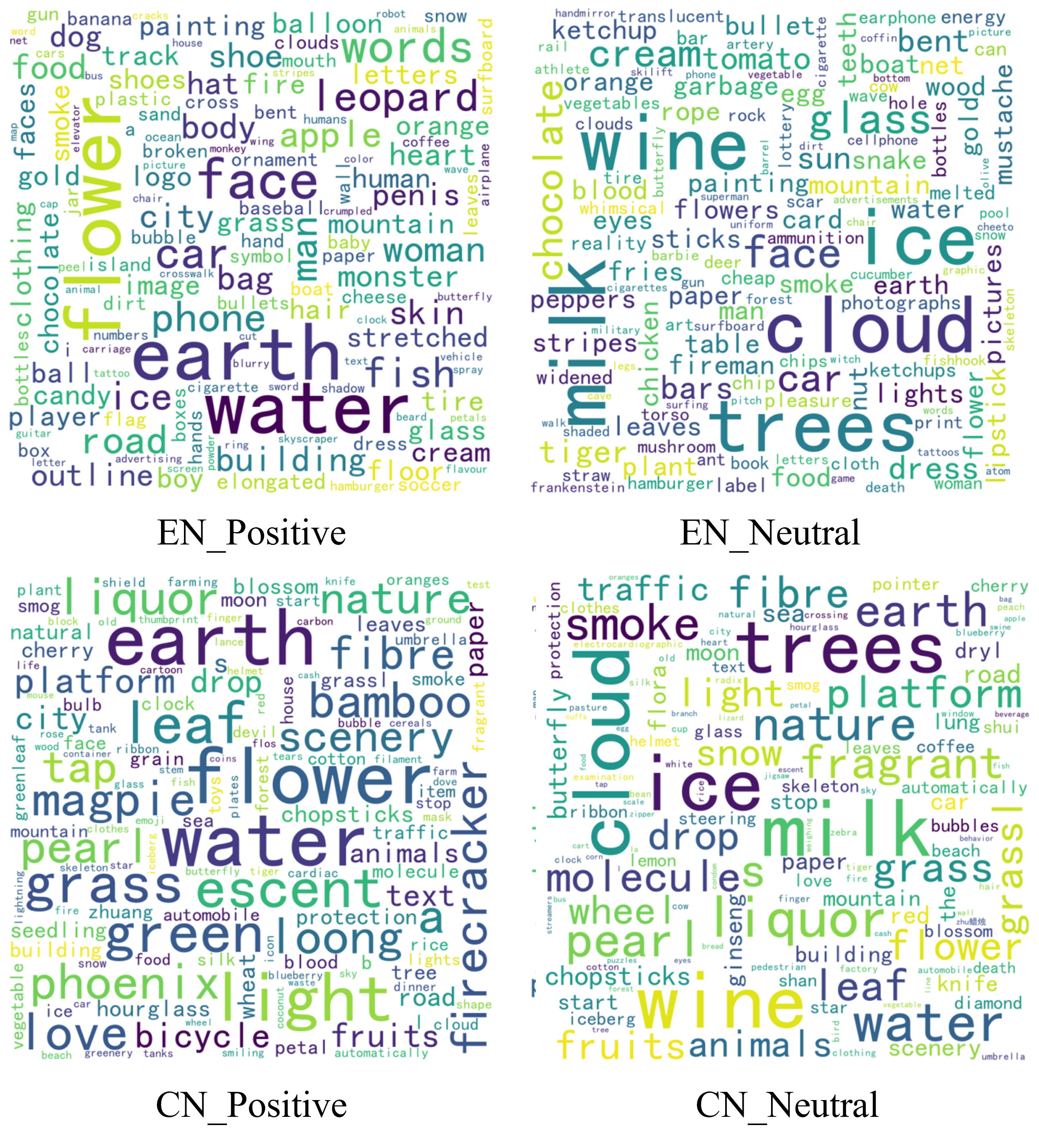}
  \caption{Word clouds of source vocabulary across sentiment categories in Chinese (CN) and English (EN) advertisements. The Chinese word cloud is translated into English for better understanding.}
  \label{fig:wordcloud}
\end{figure}

\begin{figure*}[t]
  \centering
  \includegraphics[width=0.9\linewidth]{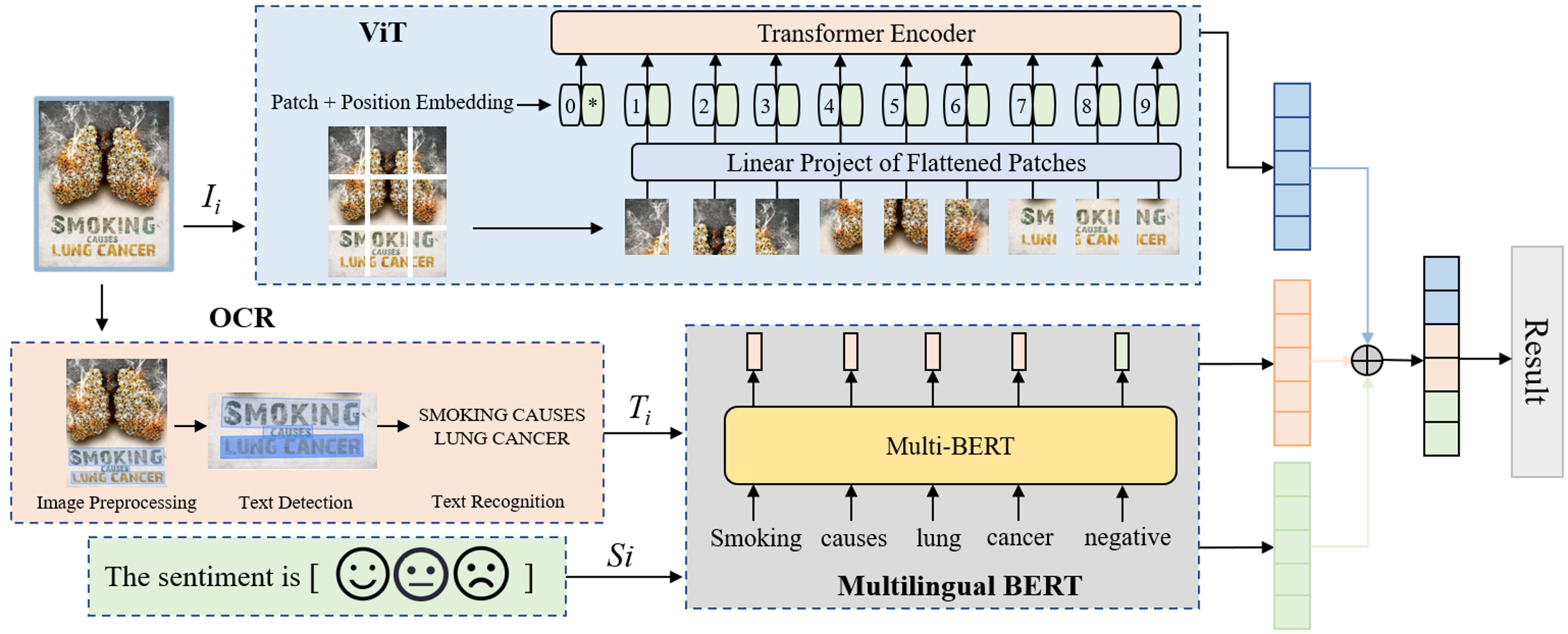}
  \caption{The SEMD framework consists of three main branches: image features extracted by ViT, text features extracted from OCR and encoded by BERT, and emotional features from sentiment analysis. These are fused for final metaphor prediction.}
  \label{fig:figure_6}
\end{figure*}

\subsection{Sentiment Vocabulary Analysis}
Our word clouds, shown in Figure~\ref{fig:wordcloud}, reveal significant overlaps between Chinese and English advertisements. Both languages frequently convey positive sentiments by utilizing words linked to universal symbols, such as `earth' (environment) and `water' (purity). These words tap into shared human experiences and cultural associations. Neutral sentiments are often conveyed through vocabulary such as `trees' in both English and Chinese advertisements, with subtle variations influenced by cultural nuances. 

However, it is important to note that the same sentiment may be expressed through different vocabulary in the two languages. For example, a metaphor used to express love or passion in English (such as `fire') may not carry the same association in Chinese, where different symbols might evoke similar emotional responses. Conversely, similar vocabulary can be employed in both languages to convey different sentiments. For instance, "storm" might be used in English to express turmoil, whereas in Chinese, the same word could evoke feelings of renewal or transformation.

\section{Methodology}
While most metaphor understanding models are developed for English datasets, their effectiveness may decline on Chinese datasets. Existing studies indicate that sentimental cognition is influenced by our shared neurophysiological structure \cite{mauro1992role,ortony2022cognitive} and is thus universal across different cultures \cite{wallbott1986universal}. Motivated by this, we develop a generalized model, Sentiment-Enriched Metaphor Detection (SEMD), which integrates sentiment information as an auxiliary feature. The model architecture is illustrated in Figure \ref{fig:figure_6}.

The model consists of three main components: two encoders, a fusion layer, and a final classifier. First, the model employs BERT \cite{devlin2019bert} as the text encoder to process textual content and sentiment information, and ViT \cite{dosovitskiy2020image} as the image encoder to process representations of text and image modalities. Both encoders output 768-dimensional feature vectors, denoted as \(T_{i}\), \(S_{i}\), and \(I_{i}\) for text, sentiment, and image, respectively. The fusion layer then applies a cascading strategy to combine these feature vectors. Once the fused vectors are obtained, a feed-forward network extracts intrinsic information from the text and images. Finally, these vectors pass through a fully connected layer, and the model generates the final results by applying the sigmoid function to the output vector.

For the metaphor detection task, our goal is to determine whether a textual-visual pair contains a metaphor. By extracting text, image, and sentiment features, concatenating them, and passing them through the feed-forward network, we obtain the prediction results after applying the activation function:
\begin{equation}
P_{Meta} = \text{Sigmoid}(\text{Fusion}(\text{concat}(I_{i},T_{i},S_{i}))).\nonumber
\end{equation}

For the sentiment analysis task, our objective is to classify the sentimental inclination of textual-visual pairs into negative, neutral, or positive sentiments. In this task, feature extraction is performed solely on text and image inputs, and the subsequent network framework is utilized to obtain the sentiment classification results:
\begin{equation}
P_{Senti} = \text{Sigmoid}(\text{Fusion}(\text{concat}(I_{i},T_{i}))).\nonumber
\end{equation}

\section{Experiments}
\subsection{Experimental Settings}
The proposed dataset, MultiMM, is evaluated using 18 existing baselines and one proposed multimodal model (i.e., SEMD). All models are built with the PyTorch framework \cite{paszke2019pytorch} and implemented using the Hugging Face library. The base networks of the model include multilingual BERT~\footnote{{\url{https://huggingface.co/bert-base-multilingual-cased}}} and ViT~\footnote{\url{https://huggingface.co/google/vit-base-patch16-224}}. We employ AdamW \cite{loshchilov2018fixing} as the optimizer and use the cross-entropy loss as the primary loss function. During training, the parameters of the pre-trained models are frozen. Model performance is evaluated based on accuracy, macro precision, and macro F1-score.

The following hyperparameters are applied for optimal performance: an embedding size of 768 to ensure robust text representation, a dropout rate of 0.3 to prevent overfitting, a maximum text length of 30 tokens to balance context capture and computational efficiency, 10 training epochs for convergence, a batch size of 64 for efficient mini-batch processing, and a learning rate ranging from 3e-5 to 5e-4 for fine-tuning weight updates.

\subsection{Baselines}
In this section, we briefly introduce the 18 baselines used in our experiment, spanning three modalities: textual (8 models), visual (3 models), and multimodal (7 models).

For the text modality, we select mBERT \cite{devlin2019bert}, Unsupervised Cross-lingual Representation Learning at Scale (XLM-R) \cite{conneau2020unsupervised}, SixTP \cite{chen2022towards}, Language-agnostic BERT Sentence Embedding (LaBSE) \cite{ansel2022composable}, GPT-3.5-TURBO \cite{gao2023scaling}, LLaMA2-8B \cite{touvron2023llama}, LLaMA3.1-70B \cite{dubey2024llama} and Deepseek-R1-70B\cite{deepseekai2025deepseekr1incentivizingreasoningcapability}. For the imaging modality, we apply VGG \cite{simonyan2014very}, ResNet \cite{he2016deep}, and ViT \cite{dosovitskiy2020image} to extract image features. For multimodal analysis, we select models designed for metaphorical language and adapt their text encoders to multilingual versions. These models include mBERT-Res \cite{zhang2021multimet}, which integrates multilingual BERT and ResNet50; Cross-Modal Graph Convolutional Networks (CMGCN) \cite{liang2022multi}, which leverage a cross-modal graph structure; Caption Enriched Samples (CES) \cite{blaier2021caption}, which incorporates image captions; MET \cite{xu2022met}, which integrates annotated vocabularies; and MFC \cite{maity2022multitask}, designed for sentiment and irony recognition. Additionally, we consider multimodal LLMs, including LLaVA~\cite{liu2023llava}, GPT-4o~\cite{hurst2024gpt}, Gemini~\cite{team2023gemini}, and Qwen2.5-VL~\cite{bai2025qwen2} models at the 3B, 7B, and 72B parameter scales.

\subsection{Performance Analysis}
\begin{table}[t]
    \centering
    \small
    \setlength{\tabcolsep}{7pt}
    \begin{tabular}{ccccc}
\toprule
    \multirow{3}{*}{{Model}} & \multicolumn{2}{c}{\makecell{{Metaphor}\\ {Detection}\\ {F1 (\%)}}} & \multicolumn{2}{c}{\makecell{{Sentiment}\\ {Analysis}\\ {F1 (\%)}}} \\ \cmidrule(lr){2-5}
        & EN & CN & EN & CN \\ \midrule
         Random & 46.70 & 41.20 & 40.41 & 35.02 \\ \midrule
        \rowcolor{cyan!20} mBERT & 64.00 & 65.52 & 62.83 & 58.43 \\ 
        \rowcolor{cyan!20} XLM\_R & \underline{69.90} & 62.38 & 69.44 & 66.21 \\
        \rowcolor{cyan!20} SixTP & 68.61 & 65.60 & 64.43 & 65.59 \\
        \rowcolor{cyan!20} LaBSE & 68.15 & \underline{66.41} & \underline{69.93} & \underline{69.02} \\ 
        \rowcolor{cyan!20} GPT-3.5-TURBO & 30.95 & 15.49 & 42.06 & 52.08 \\ 
        \rowcolor{cyan!20} Llama2-8B & 22.58 & 19.73 & 40.11 & 50.72 \\
        \rowcolor{cyan!20} Llama3.1-70B & 44.04 & 49.70 & 33.46 & 30.90 \\ 
        \rowcolor{cyan!20} Deepseek R1-70B & 44.06 & 47.95 & 53.38 & 49.01 \\ 
    \midrule

        \rowcolor{green!20} VGG16 & 70.64 & 67.85 & 59.90 & 59.02 \\ 
        \rowcolor{green!20} ResNet50 & 71.46 & 68.56 & 57.49 & 58.86 \\ 
        \rowcolor{green!20} ViT & \underline{71.67} & \underline{69.04} & \underline{61.46} & \underline{62.17} \\ 
    \midrule

        \rowcolor{orange!20} mBERT-Res & 77.83 & 74.40 & 69.75 & 68.84 \\ 
        \rowcolor{orange!20} CMGCN & \underline{79.04} & 74.91 & 72.79 & 70.19 \\ 
        \rowcolor{orange!20} CES & 78.45 & \underline{75.53} & \underline{73.24} & \underline{70.26} \\ 
        \rowcolor{orange!20} MET & 76.04 & 72.51 & 71.95 & 67.17 \\
        \rowcolor{orange!20} MFC & 75.84 & 73.25 & 71.93 & 69.60 \\ 
        \rowcolor{orange!20} GPT-4O & 64.00 & 67.00 & 36.00 & 29.00 \\ 
        \rowcolor{orange!20} LLaVA & 73.31 & 69.84 & 56.72 & 53.21 \\
        \rowcolor{orange!20} Gemini & 64.19 & 68.51 & 53.61 & 43.48 \\
        \rowcolor{orange!20} Qwen2.5-VL-3b & 57.30 & 41.69 & 52.50 & 55.26 \\
        \rowcolor{orange!20} Qwen2.5-VL-7b & 56.62 & 54.51 & 50.71 & 55.15 \\
        \rowcolor{orange!20} Qwen2.5-VL-72B & 59.12 & 67.66 & 50.44 & 57.17 \\

    \midrule

        \textbf{SEMD} & \textbf{80.16} & \textbf{77.79} & \textbf{75.69} & \textbf{70.51} \\ 

    \bottomrule
    \end{tabular}
    \caption{Experimental results on metaphor detection and sentiment analysis tasks. The best overall results are highlighted in \textbf{bold}, while the best results within each modality are \underline{underlined}. Models are categorized by modality, with textual, visual, and multimodal models highlighted in \textcolor{cyan!60}{cyan}, \textcolor{green!60}{green}, and \textcolor{orange!60}{orange}, respectively.}
    \label{tab:table_3}
\end{table}

Table \ref{tab:table_3} presents the comparative results for metaphor detection and sentiment analysis. The main text reports F1 scores, with detailed results in Appendix~\ref{sup:Performance Analysis}. The \textit{Random} entry denotes predictions made without training, using random guesses.

\textbf{Metaphor Detection.} In the textual modality, XLM-R achieves the best performance in English tasks, while LaBSE excels in Chinese tasks, benefiting from its multilingual alignment capabilities. However, GPT-3.5 TURBO and the Llama series underperform in both metaphor and sentiment recognition, particularly in Chinese tasks, where F1 scores are notably low, highlighting their limitations in handling complex metaphors. In the imaging modality, ViT outperforms ResNet50 and VGG models in both languages, demonstrating strengths in global information extraction and visual representation. The imaging modality generally surpasses the textual modality, suggesting that rich metaphorical features in advertising images enhance recognition. Among multimodal models, CMGCN and CES perform best in English and Chinese tasks, respectively, proving the effectiveness of modeling interactions between images and text. In contrast, GPT-4o, LLaVA, and Gemini exhibit relatively weak performance on multimodal tasks, particularly on Chinese tasks. Notably, Qwen2.5-VL demonstrates a corresponding improvement in metaphor detection accuracy as its parameter size increases.

\textbf{Sentiment Analysis.} In the textual modality, LaBSE leads in both Chinese (69.02\%) and English (69.93\%), possibly due to the longer average length of Chinese texts, which enhances contextual understanding. However, advanced LLMs such as GPT-3.5, Deepseek, and the Llama series perform poorly. Among them, GPT-3.5 TURBO achieves an F1 score of only 52.08\% in Chinese, DeepseekR1-70B reaches 49.01\%, Llama2-8B scores 50.72\%, and Llama3.1-70B performs even worse, with a score of just 33.46\%. In the visual modality, ViT leads in both Chinese and English, highlighting the importance of visual information in sentiment prediction. Multimodal models excel, particularly in English tasks, as CMGCN and CES achieve high F1 scores in both languages, demonstrating the effectiveness of multimodal approaches. In contrast, GPT-4o, LLaVA, Gemini, and Qwen2.5-VL struggle with sentiment analysis.

Our proposed model, SEMD, outperforms all baselines in metaphor detection and sentiment analysis by incorporating sentiment information as an auxiliary feature. As shown in Table \ref{tab:table_3}, SEMD achieves the highest F1 scores, particularly in Chinese tasks, demonstrating its robustness in cross-cultural and multimodal metaphor understanding.

\subsection{Ablation Study}
\begin{table}[t]
    \centering
    \small
    \begin{tabular}{cccc}
    \toprule
    {\makecell{{Sentiment}\\ {Feature}}} & {\makecell{{Fusion}\\ {Method}}} & {\makecell{{EN}\\ {{F1 (\%)}}}} & {\makecell{{CN}\\ {{F1 (\%)}}}} \\ \midrule
    \multirow{3}{*}{w/o} & add & 75.95 & 72.05 \\
     & max & 76.29 & 74.09 \\
     & concat & \underline{78.64} & \underline{74.84} \\ \midrule
    \multirow{3}{*}{w/} & add & 77.76 & 74.37 \\
     & max & 78.59 & 74.37 \\
     & concat & \textbf{80.88} & \textbf{77.39} \\ \bottomrule
    \end{tabular}
    \caption{Results of the ablation study. The best and second-best results are highlighted in \textbf{bold} and \underline{underline}, respectively.}
    \label{tab:table_5}
\end{table}

To evaluate the impact of fusion methods and sentiment features, we conduct an ablation study under metaphor detection tasks. We compare three fusion methods (add, max, and concat) and test the effects of removing sentiment features. As shown in Table \ref{tab:table_5}, the concat fusion method effectively integrates text and image features, achieving the best performance across both Chinese and English datasets. Furthermore, including sentiment features significantly enhances recognition performance compared to their absence. Note that the proposed SEMD uses the concat fusion method with sentiment features, which corresponds to the last entry in the table.

\subsection{Metaphor Detection via Bidirectional Chinese-English Translation}
\begin{table}[t]
    \centering
    \small
    \setlength{\tabcolsep}{4pt}
    \begin{tabular}{cc>{\columncolor{cyan!20}}c>{\columncolor{orange!20}}cc}
    \toprule
     &  & mBERT & mBERT-Res & SEMD \\ \midrule
    \multirow{3}{*}{EN $\to$ CN} & Acc (\%) & 64.29 & 76.35 & 78.57 \\
     & Pre (\%) & 63.72 & 76.73 & 78.19 \\
     & F1 (\%) & 59.70 & 75.63 & 77.64 \\ \midrule
    \multirow{3}{*}{CN $\to$ EN} & Acc (\%) & 65.22 & 73.86 & 75.45 \\
     & Pre (\%) & 67.46 & 75.90 & 77.73 \\
     & F1 (\%) & 65.31 & 73.96 & 75.53 \\ \bottomrule
    \end{tabular}
    \caption{The results of metaphor detection tasks after Chinese-English translation. Experimental results on metaphor detection task via bidirectional Chinese-English translation.}
    \label{Translation}
\end{table}

To explore issues related to language understanding and expression in cross-cultural communication, we conduct supplementary experiments using Chinese-English translated texts for metaphor detection tasks. We evaluate the performance of the pure text model mBERT, the multimodal model mBERT-Res, and our proposed model on this task. As shown in Table~\ref{Translation}, the results indicate a performance drop in metaphor detection when using translated texts compared to the original versions. We attribute this decline to two main factors.

First, significant differences in language structure and grammar often lead to the loss of semantic and rhetorical nuances of metaphors during direct translation, thereby reducing the accuracy of metaphor detection.

Second, the understanding and expression of metaphors heavily rely on cultural background and contextual factors. Variations in values, belief systems, customs, and historical backgrounds across cultures result in different interpretations and perceptions of the same metaphor. As a result, simple language conversion often fails to fully capture and reproduce the subtle cultural and pragmatic meanings embedded in metaphors.

\subsection{Error Analysis}
\begin{CJK*}{UTF8}{gbsn} 
We conduct an error analysis of the test set prediction results for metaphor detection and sentiment analysis tasks. The error analysis of metaphor detection shows that the accuracy of Chinese metaphor detection is 73\%, lower than the 86\% accuracy for English. This discrepancy may result from cultural and ideological influences in the advertising data. Chinese metaphors tend to be more subtle, while English metaphors are more straightforward. The error analysis of sentiment analysis reveals that although the Chinese dataset has a higher proportion of neutral sentiment, its recognition accuracy is only 55\%, compared to 62\% for English. The low performance in Chinese is due to lexical ambiguity and polysemy. Many misclassified cases contain positive words (e.g., cherish~[珍爱], civilization~[文明]) but lack an overall positive sentiment, leading to errors. Due to page limitations, detailed results and discussion are provided in Appendix~\ref{error}. 

In addition, to vividly illustrate the profound impact of cultural context on metaphor understanding, we also provides a case study in Appendix~\ref{CaseStudy}.
\end{CJK*}

\section{Conclusion}

Metaphors play a fundamental role in communication, yet NLP research predominantly relies on English datasets, introducing cultural biases that limit cross-cultural applicability. To address this, we introduce MultiMM, a benchmark dataset designed for multimodal metaphor studies in Chinese and English. MultiMM consists of 8,461 text-image advertisement pairs with annotations, enabling a more comprehensive evaluation of metaphor processing across cultures. We also propose SEMD, a sentiment-enriched model that integrates sentiment embeddings to enhance multilingual and multimodal metaphor comprehension. Experimental results highlight the significant impact of cultural bias on metaphor detection and sentiment analysis. SEMD outperforms all baselines, demonstrating its robustness in cross-cultural and multimodal metaphor understanding. We hope MultiMM serves as a valuable resource for multicultural multimodal metaphor processing and that SEMD offers insights for improving cross-cultural NLP models.

\section*{Limitations}
While MultiMM is a valuable resource for studying multimodal metaphors, it has limitations that present opportunities for future research. First, the dataset is currently limited to advertising, as annotating multimodal metaphors across genres (e.g., social media, news, literature) poses challenges in consistency and scalability. Expanding to these domains would enable a more comprehensive analysis of multimodal metaphors in diverse contexts.

Second, our work focuses on English and Chinese, representing Western and Eastern cultures, respectively. While this provides a foundational comparison, it does not capture the full diversity of global languages and cultures. Future research could extend MultiMM to more languages and cultural frameworks, deepening insights into how multimodal metaphors vary across linguistic and cultural landscapes.
\section*{Acknowledgments}
We would like to thank the anonymous reviewers for their insightful and valuable comments.
This work was supported by the Key Program of the National Language Commission (No. ZDI145-80) and the National Natural Science Foundation of China (NSFC) (No. 62076051).



\bibliography{custom}

\appendix

\section{Supplementary Experimental Results}
\subsection{Performance Analysis}
\label{sup:Performance Analysis}
In this section, we provide detailed results for metaphor detection and sentiment analysis on 18 baselines. We use accuracy (Acc), precision (Pre), and F1 score (F1) as metrics to compare performance.

\begin{table*}[t]
    \centering
    \setlength{\tabcolsep}{6pt}
    \begin{tabular}{ccccccc}
    \toprule
     & \multicolumn{3}{c}{\textbf{EN}}  & \multicolumn{3}{c}{\textbf{CN}} \\ 
    \midrule
    
     Model & Acc (\%) & Pre (\%) & F1 (\%) & Acc (\%) & Pre (\%) & F1 (\%) \\ 

    \midrule
     Random & 50.25 & 46.21 & 46.70 & 45.95 & 52.88 & 41.20 \\ 
    \midrule
        \rowcolor{cyan!20} mBERT & 66.50 & 66.89 & 64.00 & 65.65 & 65.44 & 65.52 \\ 
        \rowcolor{cyan!20} XLM\_R & \underline{69.70} & \underline{70.53} & \underline{69.90} & 65.43 & 65.84 & 62.38 \\
        \rowcolor{cyan!20} SixTP & 68.71 & 68.55 & 68.61 & 66.74 & 66.30 & 65.60 \\
        \rowcolor{cyan!20} LaBSE & 68.72 & 68.31 & 68.15 & \underline{66.96} &66.49 & \underline{66.41} \\ 
        \rowcolor{cyan!20} GPT-3.5-TURBO & 42.85 & 52.00  & 30.95 & 45.45 & \underline{68.75} & 15.49\\ 
        \rowcolor{cyan!20} Llama2-8B & 40.74 & 47.29  & 22.58 & 44.54 & 57.69  & 19.73 \\
        \rowcolor{cyan!20} Llama3.1-70B & 45.18 &  53.40  & 44.04& 49.70 & 57.80 & 49.70 \\
        \rowcolor{cyan!20} Deepseek R1-70B & 45.00 &  52.81  & 44.06& 52.00 & 61.34 & 47.95 \\

    \midrule
        \rowcolor{green!20} VGG16 & 70.44 & 71.60 & 70.64 & 68.27 & 67.84 & 67.85 \\ 
        \rowcolor{green!20} ResNet50 & 72.91 & 73.92 & 71.46 & 71.55 & 75.35 & 68.56 \\ 
        \rowcolor{green!20} ViT & \underline{73.40} & \underline{75.14} & \underline{71.67} & \underline{71.99} & \underline{75.99} & \underline{69.04} \\ 

    \midrule
        \rowcolor{orange!20} mBERT-Res & 77.83 & \underline{80.61} & 77.83 & 75.49 & 76.46 & 74.40 \\ 
        \rowcolor{orange!20} CMGCN & \textbf{79.56} & 79.96 & \underline{79.04} & 75.93 & 76.65 & 74.91 \\ 
        \rowcolor{orange!20} CES & 78.57 & 78.46 & 78.45 & \underline{76.15} & 76.24 & \underline{75.53} \\ 
        \rowcolor{orange!20} MET & 76.15 & 76.01 & 76.04 & 74.18 & 75.74 & 72.51 \\
        \rowcolor{orange!20} MFC & 77.34 & 79.97 & 75.84 & 74.84 & \underline{76.74} & 73.25 \\ 
        \rowcolor{orange!20} GPT-4O & 54.00 & 63.00 & 64.00  & 57.00&  62.00&67.00   \\ 
        \rowcolor{orange!20} LLaVA & 59.11 & 59.06  & 73.31 & 56.81 & 58.20  & 69.84 \\
        \rowcolor{orange!20} Gemini &64.75   &64.19    & 64.19 &69.00  &68.72    & 68.15 \\
        \rowcolor{orange!20}Qwen2.5-VL-3b &56.5   &59.01    & 57.30 &32.78   & 32.78   & 41.69 \\
        \rowcolor{orange!20}Qwen2.5-VL-7b&63.00  &60.08   &     56.62 &63.00 &72.14  & 54.51 \\
        \rowcolor{orange!20}Qwen2.5-VL-72b&60.00  & 58.83  & 59.12   & 70.25&72.95    & 67.66 \\      
    \midrule
        \textbf{SEMD} & \underline{79.50} & \textbf{82.72} & \textbf{80.16} & \textbf{77.75} & \textbf{77.84} & \textbf{77.79} \\ 
    \bottomrule
    \end{tabular}
        \caption{Experimental results on metaphor detection task. The best overall results are highlighted in \textbf{bold}, while the best results within each modality are \underline{underlined}. Models are categorized by modality, with textual, visual, and multimodal models highlighted in \textcolor{cyan!60}{cyan}, \textcolor{green!60}{green}, and \textcolor{orange!60}{orange}, respectively.}
    \label{tab:table_3_1}
\end{table*}

\textbf{Metaphor Detection.} We present the detailed results of metaphor detection in Table \ref{tab:table_3_1}. The Random entry denotes predictions made by the model without training, relying purely on random guessing. Overall, multimodal approaches significantly outperform both text-only and image-only models, demonstrating their ability to leverage complementary information from both modalities. Additionally, English datasets yield better results than their Chinese counterparts, likely due to the abundance of pre-trained English-language data, which facilitates model learning.

Among text-based models, XLM-R achieves the highest performance for English, while LaBSE demonstrates superior results for Chinese, highlighting its robust multilingual alignment capabilities. The strong performance of these models underscores the importance of cross-lingual knowledge transfer in metaphor comprehension.

In the visual domain, ViT outperforms both ResNet50 and VGG models in both languages, reinforcing the superiority of Transformer-based architectures in capturing global image features. The generally stronger performance of image models over text models suggests that advertising images contain rich metaphorical elements, which models can effectively leverage for metaphor detection.

Among multimodal approaches, CMGCN and CES stand out as top performers for English and Chinese, respectively. CMGCN excels at fine-grained feature extraction from images, while CES enhances metaphor comprehension by incorporating image captions as global auxiliary features, effectively modeling the interplay between text and images. Notably, mBERT-Res, which replaces VGG with ResNet, outperforms MET, highlighting the critical role of strong visual feature extraction in improving overall performance. In contrast, MFC, a multitask model jointly trained on metaphor and sentiment recognition, struggles to achieve competitive results, likely due to interference between the two learning objectives.

Our proposed model, SEMD, achieves the best overall results, validating the effectiveness of sentiment-aware approaches in multimodal metaphor processing.

\begin{table*}[t]
    \centering
    \setlength{\tabcolsep}{6pt}
    \begin{tabular}{ccccccc}
    \toprule
     & \multicolumn{3}{c}{\textbf{EN}}  & \multicolumn{3}{c}{\textbf{CN}} \\ 
    \midrule
    Model & Acc (\%) & Pre (\%) & F1 (\%) & Acc (\%) & Pre (\%) & F1 (\%) \\ 
    \midrule
     Random & 39.41 & 45.82 & 40.41 & 34.48 & 41.47 & 35.02 \\ 
    \midrule
        \rowcolor{cyan!20} mBERT & 70.27 & 68.97 & 62.83 & 67.76 & 64.26 & 58.43 \\ 
        \rowcolor{cyan!20} XLM\_R & \underline{71.50} & 69.34 & 69.44 & 68.71 & 65.00 & 66.21 \\
        \rowcolor{cyan!20} SixTP & 69.46 & 66.50 & 64.43 & 68.49 & 64.53 & 65.59 \\
        \rowcolor{cyan!20} LaBSE & 70.44 & \underline{69.44} & \underline{69.93} & \underline{70.02} & \underline{68.04} & \underline{69.02} \\ 
        \rowcolor{cyan!20} GPT-3.5-TURBO & 42.36 & 56.7 & 42.06 & 51.36 & 52.91 & 52.08\\ 
        \rowcolor{cyan!20} Llama2-8B & 40.64 & 54.49 & 40.11 & 50.22 & 51.39 & 50.72 \\
        \rowcolor{cyan!20} Llama3.1-70B &  52.49 & 34.61 &33.46  & 51.00 &  37.90  &30.90 \\
        \rowcolor{cyan!20} Deepseek R1-70B &53.50 &  54.25  & 53.83& 53.50 & 49.90 & 49.01 \\
    \midrule
        \rowcolor{green!20} VGG16 & 65.60 & 59.66 & 59.90 & 65.52 & 58.66 & 59.02 \\ 
        \rowcolor{green!20} ResNet50 & 68.55 & \underline{66.70} & 57.49 & 68.49 & \underline{66.19} & 58.86 \\ 
        \rowcolor{green!20} ViT & \underline{69.04} & 65.84 & \underline{61.46} & \underline{68.64} & 65.55 & \underline{62.17} \\ 
 
    \midrule
        \rowcolor{orange!20} mBERT-Res & 73.46 & 71.58 & 69.75 & 72.21 & 68.95 & 68.84 \\ 
        \rowcolor{orange!20} CMGCN & 74.40 & 71.59 & 72.79 & 73.03 & \textbf{70.84} & 70.19 \\ 
        \rowcolor{orange!20} CES & \underline{74.56} & \underline{72.52} & \underline{73.24} & \underline{73.30} & 69.76 & \underline{70.26} \\ 
        \rowcolor{orange!20} MET & 73.68 & 71.46 & 71.95 & 71.55 & 68.06 & 67.17 \\
        \rowcolor{orange!20} MFC & 73.30 & 70.76 & 71.93 & 72.43 & 68.91 & 69.60 \\ 
        \rowcolor{orange!20} GPT-4O & 61.00 &40.00    &36.00   &50.00&39.00    &29.00   \\ 
        \rowcolor{orange!20} LLaVA & 55.41 & 58.61 & 56.72 & 52.95 & 53.59 & 53.21 \\
        \rowcolor{orange!20} Gemini &51.75   &56.61    & 53.61 & 37.75 & 53.24   & 43.48 \\
        \rowcolor{orange!20}Qwen2.5-VL-3b &56.00   &53.11    & 52.50 & 60.25  & 63.51   & 55.26 \\
        \rowcolor{orange!20}Qwen2.5-VL-7b&59.00  &59.81   &     50.17 & 60.75&58.85  & 55.15 \\
        \rowcolor{orange!20}Qwen2.5-VL-72b&53.75  &51.39   & 50.44   & 55.75&64.33    & 57.17 \\      
        
    \midrule
        \textbf{SEMD} & \textbf{76.60} & \textbf{74.83} & \textbf{75.69} & \textbf{73.40} & \underline{70.66} & \textbf{70.51} \\ 
  
    \bottomrule
    \end{tabular}
        \caption{Experimental results on sentiment analysis task. The best overall results are highlighted in \textbf{bold}, while the best results within each modality are \underline{underlined}. Models are categorized by modality, with textual, visual, and multimodal models highlighted in \textcolor{cyan!60}{cyan}, \textcolor{green!60}{green}, and \textcolor{orange!60}{orange}, respectively.}
    \label{tab:table_4}
\end{table*}

\textbf{Sentiment Analysis.}  Table \ref{tab:table_4} presents the results of sentiment recognition on the test set. As in metaphor detection tasks, multimodal approaches generally outperform unimodal ones, and English results surpass those in Chinese. However, unlike metaphor detection, the text modality outperforms the image modality in sentiment recognition. This is likely because textual information more directly conveys sentiment.

Within the text modality, although LaBSE achieves higher precision and F1 scores on English data, XLM-R still attains higher accuracy. LaBSE also yields the best results on Chinese textual data. In the image modality, ViT consistently outperforms VGG and ResNet. Among multimodal models, CES achieves the highest accuracy in both Chinese and English, suggesting that image captions are particularly helpful for sentiment recognition. This may be because captions directly describe the overall advertisement, providing more clues about sentiment and theme.

Notably, the MET model shows improvement in both English and Chinese when introducing source and target domains as auxiliary features, outperforming mBERT-Res and highlighting its effectiveness. Our proposed model, SEMD, achieves the best results in both English and Chinese, further demonstrating the benefits of incorporating sentiment features.



    
      
        
        
        

\begin{table}[t]
    \centering
    \begin{tabular}{ccc}
        \toprule
        \textbf{Task}  & {\makecell{{EN}\\ {{Acc (\%)}}}} & {\makecell{{CN}\\ {{Acc (\%)}}}}\\ 
        \midrule
        Sentiment Analysis & 62.00 & 55.00 \\ 
        Metaphor Detection & 86.00 & 73.00 \\ 
        \bottomrule
    \end{tabular}
    \caption{The accuracy differences between Chinese and English data in sentiment analysis and metaphor detection tasks.}
    \label{tab_neutral} 
\end{table}

\begin{CJK*}{UTF8}{gbsn} 
\begin{figure*}[t]
    \centering
    \subfigure[Don't fill your closet with cruelty. Wear VEGAN.]{
        \includegraphics[width=0.25\linewidth]{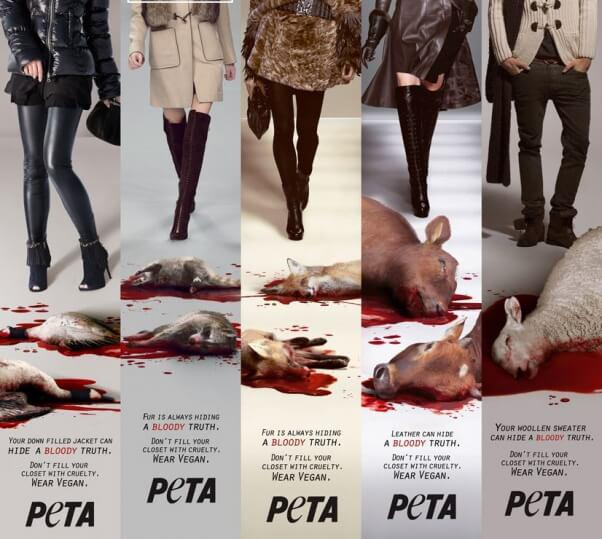}
        \label{Fig:001}
    }
    \hspace{0.25em}
    \subfigure[Smoking kills.]{
        \includegraphics[width=0.3\linewidth]{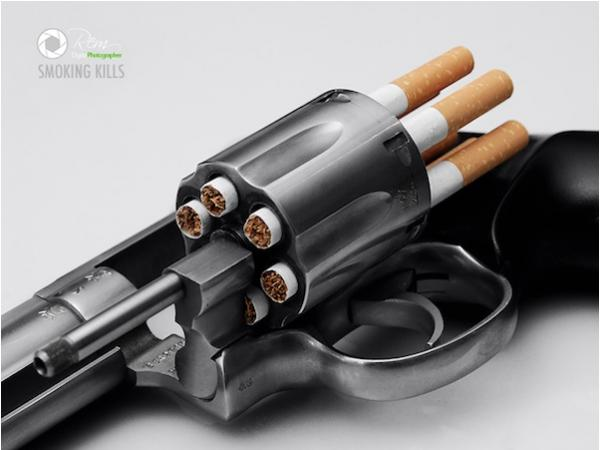}
        \label{Fig:002}
    }
    \hspace{0.25em}
     \subfigure[Words can kill. Say no to discrimination. We are all the same.]{
        \includegraphics[width=0.35\linewidth]{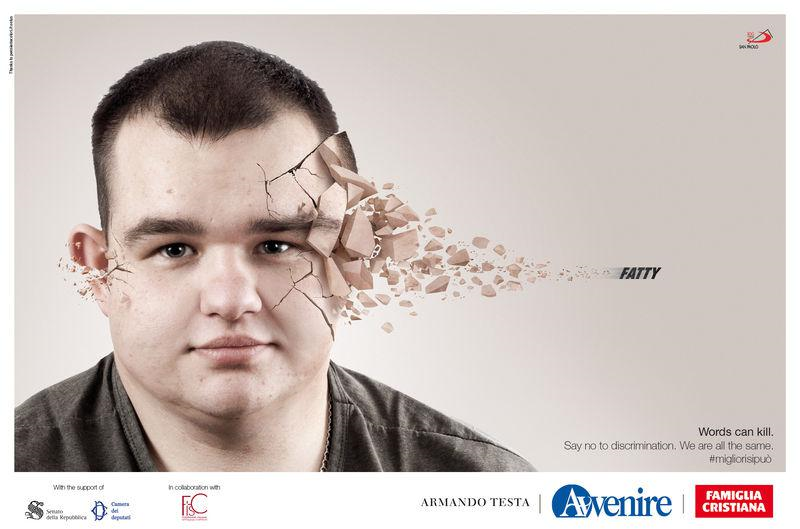}
        \label{Fig:003}
    }
    \hspace{0.25em}
     \subfigure[爱是拒绝, 爱是责任 ~(Love is rejection, love is responsibility.)]{
        \includegraphics[width=0.22\linewidth]{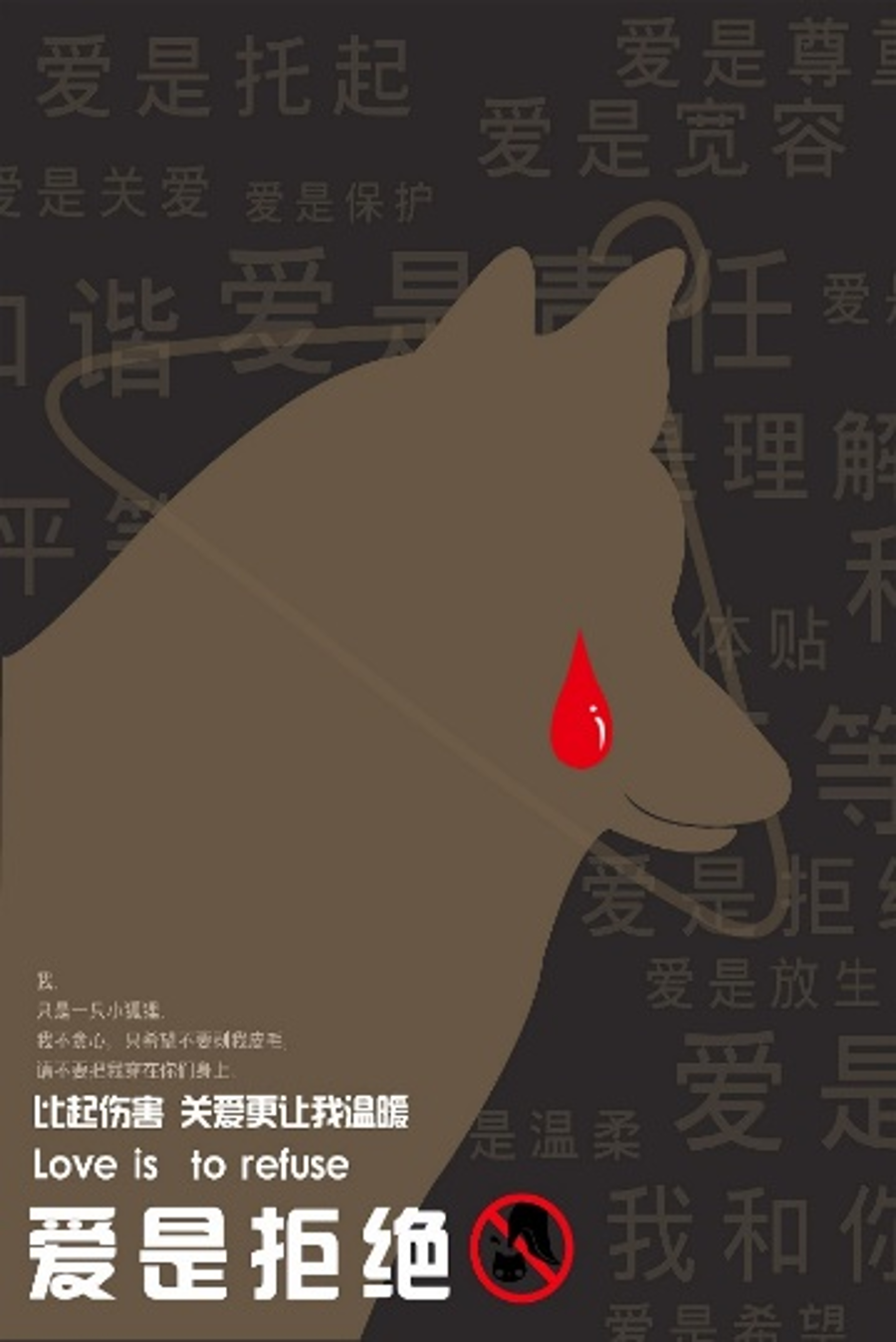}
        \label{Fig:004}
    }
    \hspace{0.25em}
     \subfigure[吸烟有害健康  ~(Smoking is harmful to health.)
]{
        \includegraphics[width=0.22\linewidth]{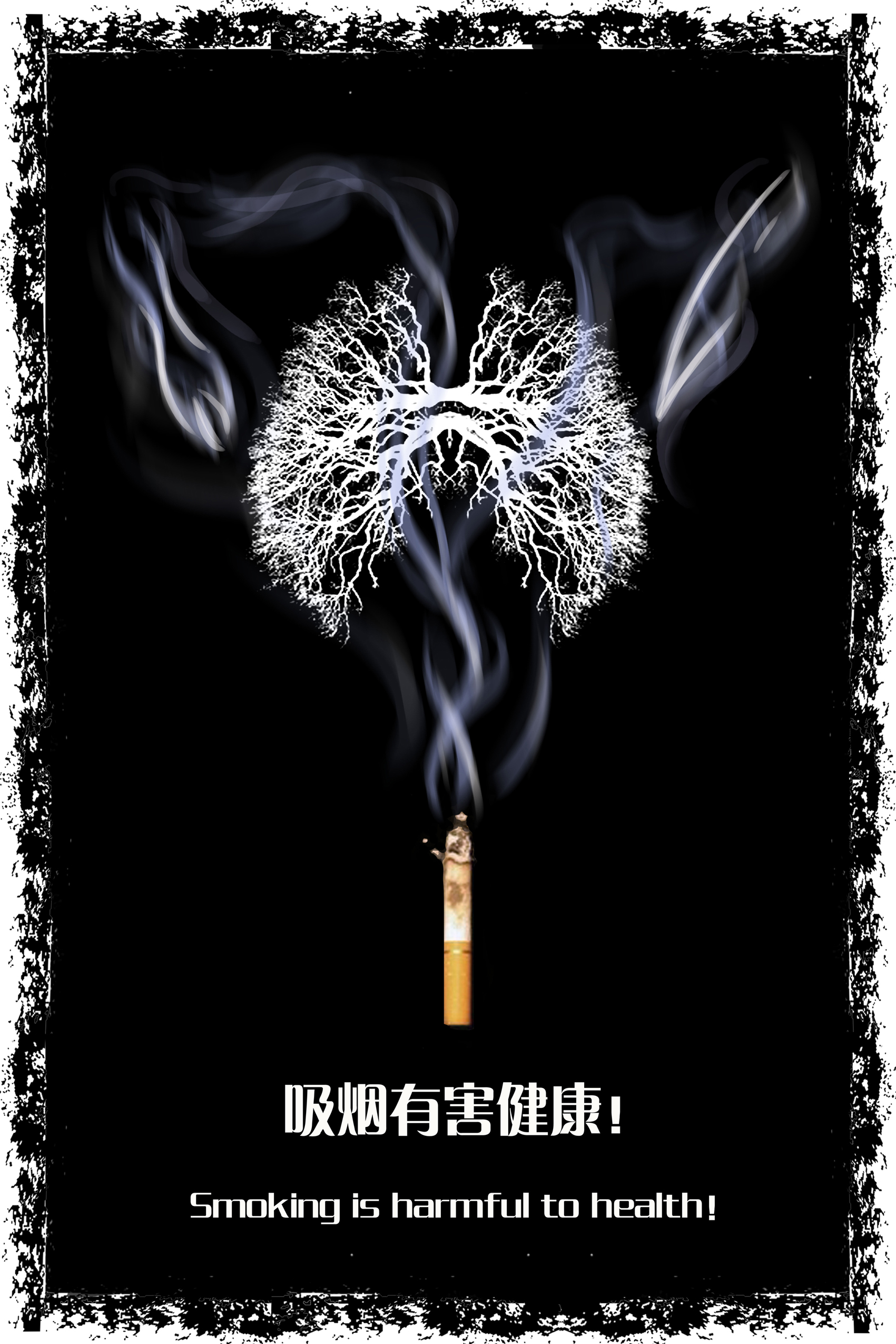}
        \label{Fig:005}
    }
    \hspace{0.25em}
     \subfigure[拒绝网络暴力, 拒绝键盘侠  ~(Reject online violence, reject keyboard warriors.)]{
        \includegraphics[width=0.23\linewidth]{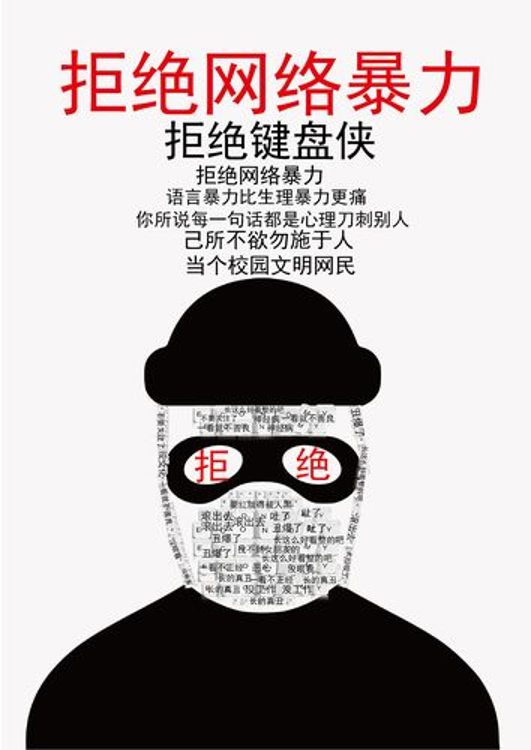}
        \label{Fig:006}
    }
    \hspace{0.25em}
     \subfigure[共创美好安徽  ~(Let's work together to create a better Anhui.)]{
        \includegraphics[width=0.22\linewidth]{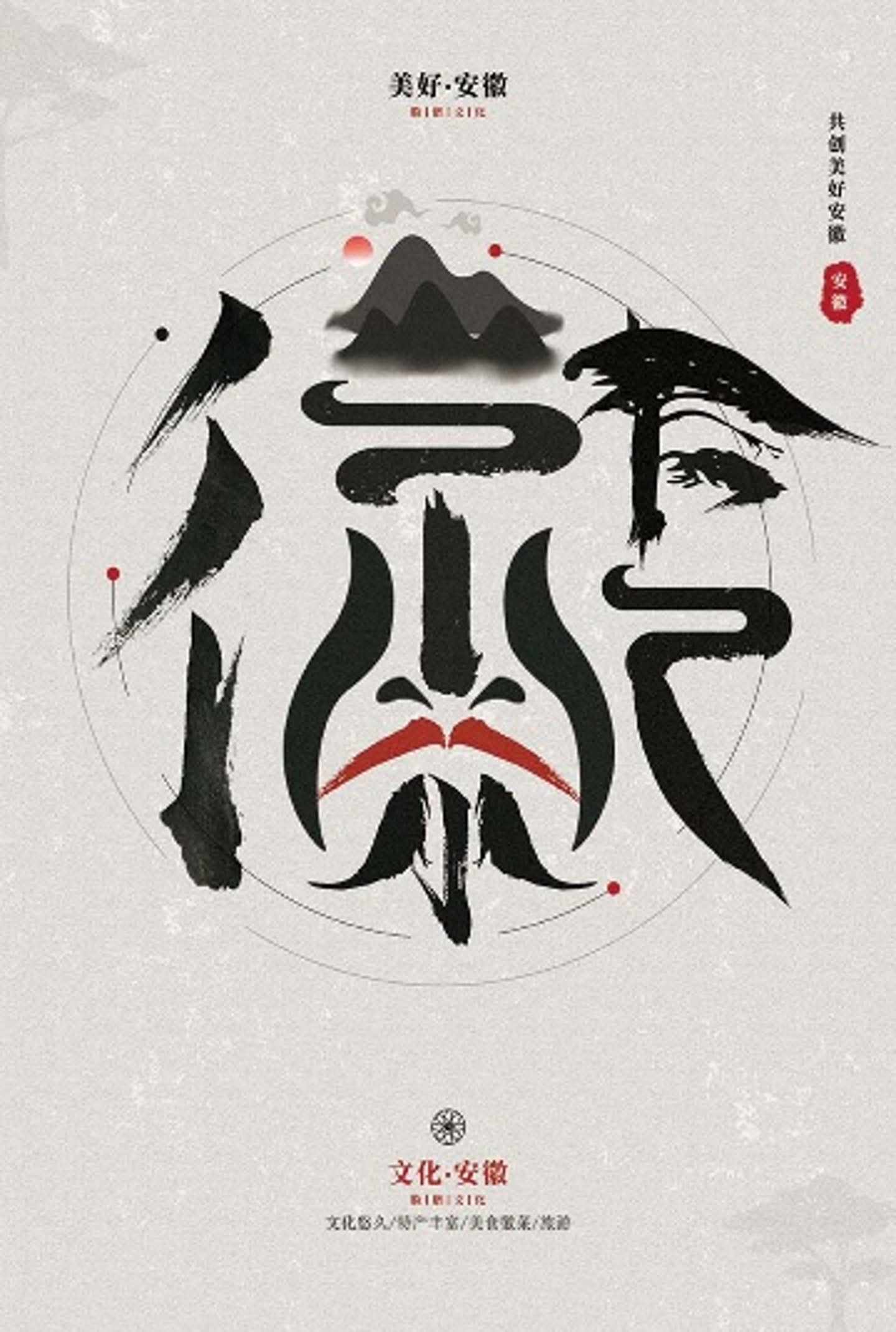}
        \label{Fig:007}
    }
    \caption{Examples of metaphors.} 
    \label{fig:figure_8}
\end{figure*}
\end{CJK*}

\subsection{Error Analysis}
\label{error}

In analyzing the results of the metaphor task, as shown in Table \ref{tab_neutral}, we examine the accuracy of metaphorical samples and observe a notable disparity between English and Chinese metaphorical image-text recognition. The accuracy for Chinese is only 73\%, whereas English data achieves 86\%. The lower accuracy in Chinese metaphor recognition is likely due to cultural and ideological influences in advertising data. Specifically, Chinese metaphors tend to be more implicit, while in English advertising, metaphors are often more direct.

Figure \ref{fig:figure_8} provides examples of shared metaphor themes in Chinese and English data. For instance, Figures \ref{Fig:001} and \ref{Fig:004} depict metaphors related to animal protection. In English data, the comparison between deceased animals and clothing implies that leather garments contribute to animal deaths. In contrast, Chinese data conveys harm to animals through imagery, where tears resembling blood flow from animal figures, while text in the image expresses the concept of love and protection.

Figures \ref{Fig:002} and \ref{Fig:005} revolve around the theme of smoking as harmful to health. English data depicts cigarettes as bullets, signifying that smoking can lead to death. In contrast, Chinese data subtly conveys the harm of smoking by shaping smoke patterns to resemble lungs, emphasizing its impact on lung health.

Figures \ref{Fig:003} and \ref{Fig:006} illustrate themes of verbal harm and discrimination. English data depicts the damage of language through fractured facial expressions and bullet trajectories, whereas Chinese data uses harmful words to form the image of a "bad person," criticizing verbal attacks.

\begin{CJK*}{UTF8}{gbsn} 
Furthermore, Chinese data tends to rely more on cultural connotations, such as family values, collectivism, and traditional culture. In Figure \ref{Fig:007}, local culture is promoted by incorporating drama masks and landscapes to form the character `徽' (an abbreviation for Anhui Province in China), which frequently appears in Chinese advertisements. In conclusion, cultural and linguistic differences increase the difficulty and complexity of Chinese metaphor detection, making it more challenging for models to distinguish metaphorical data.
\end{CJK*}

\begin{CJK*}{UTF8}{gbsn} 

In the sentiment analysis task, we observe that in the Chinese dataset, despite a higher proportion of neutral sentiment compared to the English dataset, the accuracy of neutral sentiment recognition is relatively low, at only 55\%. In contrast, English data achieves an accuracy of 62\%. An analysis of the results reveals that the pure text modality significantly outperforms the image modality, with text playing a predominant role in sentiment recognition tasks.

We examine cases where Chinese predictions of neutral sentiment are incorrect and identify a common pattern. Specifically, these instances contain words with positive sentiment (e.g., `珍爱' (cherish), `文明' (civilization), `珍视' (treasure)), yet the overall message does not express positive sentiment. Clearly, these words lead to model misjudgments. This relates to the polysemy and ambiguity of Chinese vocabulary, where the same word can carry entirely different meanings depending on context. For example, in Chinese, `好" (good) is typically positive, but in certain contexts, it can convey negative sentiment, as in `好烦' (very annoying). Moreover, Chinese grammatical structures and particles significantly impact sentiment interpretation, further affecting the model’s ability to accurately recognize sentiment.

Due to the limited proportion of negative sentiment in the dataset, the model lacks sufficient negative samples during training, limiting its ability to recognize and capture negative sentiment. Therefore, the challenges in multimodal sentiment recognition primarily stem from the scarcity of negative sentiment data, the polysemy of neutral sentiment, and cultural influences. Future research can explore strategies to address these challenges and enhance sentiment recognition across diverse cultural datasets.

\section{Case Study}
\label{CaseStudy}

\begin{figure}[t]
    \centering
    \subfigure[Ashes to Ashes.]{
        \includegraphics[width=0.325\linewidth]{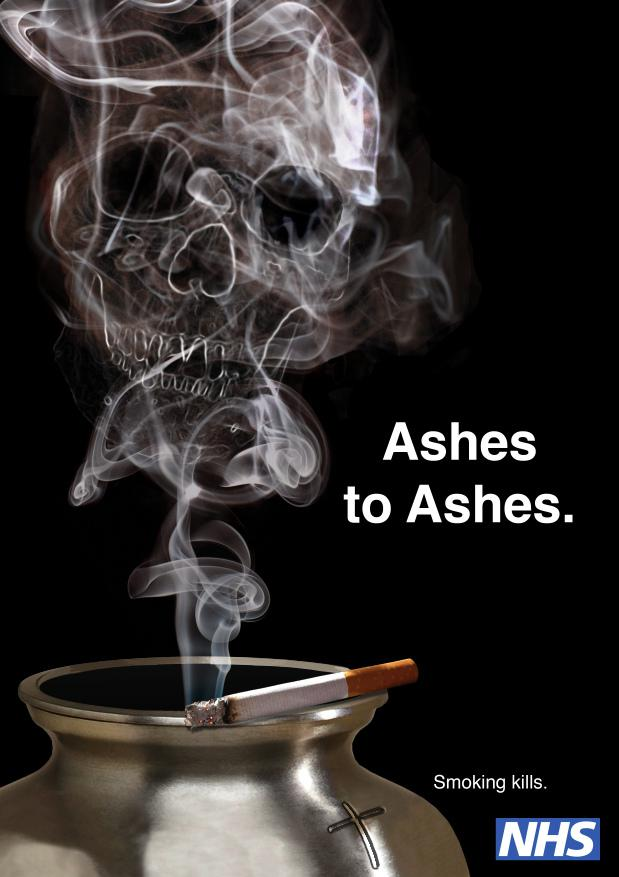}
        \label{casestudy1}
    }
    \hspace{0.25em}
    \subfigure[禁止吸烟 ~(No smoking.)]{
        \includegraphics[width=0.5\linewidth]{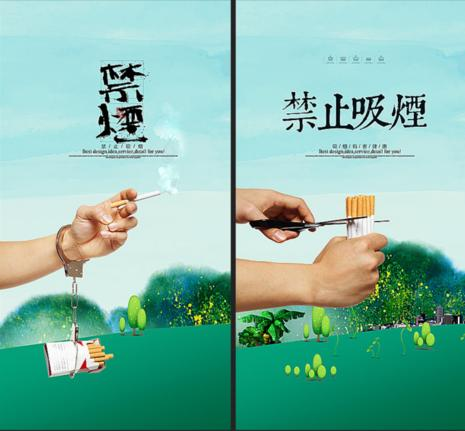}
        \label{casestudy2}
    }
    \caption{Metaphor comparison across cultures for a common theme.} 
\end{figure}

Cultural bias significantly affects model performance in cross-cultural metaphor understanding. Western advertisements often use shocking visuals to convey anti-smoking messages. For example, the `Ashes to Ashes' advertisement in Figure \ref{casestudy1} features a visual metaphor where smoke forms a skull, creating a direct and intense visual impact that evokes negative emotions like fear and despair. These advertisements clearly illustrate the deadly consequences of smoking through a combination of images and text, making it easier for the model to recognize the negative sentiment.

In contrast, Chinese advertisements present anti-smoking messages more subtly, often using warning symbols instead of shocking imagery. For instance, Figure \ref{casestudy2} depicts `handcuffs' and `scissors' to symbolize restraint and quitting smoking, emphasizing caution rather than shock. While this effectively conveys an anti-smoking message, its emotional tone is milder, making it more challenging for models to capture the emotional intensity from the image alone. As a result, sentiment classification models tend to perform worse on Chinese advertisements compared to English advertisements.

These examples highlight the profound influence of cultural context on metaphor understanding. Western culture favors direct and emotionally intense expressions, while Chinese advertisements rely on cautious symbols and milder emotional tones. Consequently, models face challenges in accurately interpreting metaphors across such cultural differences.

\end{CJK*}

\end{document}